\documentclass[lettersize,journal]{IEEEtran}
\usepackage{amsmath,amsfonts}
\usepackage{algorithmic}
\usepackage{algorithm}
\usepackage{array}
\usepackage[caption=false,font=normalsize,labelfont=sf,textfont=sf]{subfig}
\usepackage{textcomp}
\usepackage{stfloats}
\usepackage{url}
\usepackage{verbatim}
\usepackage{graphicx}
\usepackage{cite}
\usepackage{microtype}
\usepackage{paralist}
\usepackage{booktabs} %
\usepackage{hyperref}
\usepackage{mathrsfs}
\usepackage{amssymb}
\usepackage{mathtools}
\usepackage{amsthm}
\usepackage[capitalize,noabbrev]{cleveref}
\usepackage[textsize=tiny]{todonotes}
\theoremstyle{plain}
\newtheorem{theorem}{Theorem}[section]

\theoremstyle{definition}
\newtheorem{definition}[theorem]{Definition}

\theoremstyle{remark}

\newcommand{\BEST}[1]{\textcolor{red}{\bf #1}}
\newcommand{\SECOND}[1]{\textcolor{blue}{\bf #1}}
\newcommand{\THIRD}[1]{\textcolor{violet}{\bf #1}}

\begin{document}

\title{Graph Neural Aggregation-diffusion with Metastability}

\author{Kaiyuan Cui\textsuperscript{*}\thanks{Kaiyuan Cui is with the Academy of Mathematics and Systems Science, Chinese Academy of Sciences, Chinese Academy of Sciences, Beijing, China. Email:cuiky@amss.ac.cn}, Xinyan Wang\textsuperscript{*}\thanks{Xinyan Wang is a doctoral student at the Academy of Mathematics and Systems Science, Chinese Academy of Sciences, Chinese Academy of Sciences, Beijing, China. Email:wangxinyan@amss.ac.cn}, Zicheng Zhang\textsuperscript{*}\thanks{Zicheng Zhang is a doctoral student at University of Chinese Academy of Sciences, Beijing, China. Email: zhangzicheng19@mails.ucas.ac.cn.}, Weichen Zhao\textsuperscript{*}\thanks{Weichen Zhao is with the School of Statistics and Data Science, Nankai University, Tianjin, China. Email: zhaoweichen@nankai.edu.cn}}

\maketitle
\renewcommand{\thefootnote}{\fnsymbol{footnote}}
\footnotetext[1]{Authors are listed in aphabetic order.}
\begin{abstract}
    Continuous graph neural models based on differential equations have expanded the architecture of graph neural networks (GNNs). Due to the connection between graph diffusion and message passing, diffusion-based models have been widely studied. However, diffusion naturally drives the system towards an equilibrium state, leading to issues like over-smoothing. To this end, we propose GRADE inspired by \underline{GR}aph \underline{A}ggregation-\underline{D}iffusion \underline{E}quations, which includes the delicate balance between nonlinear diffusion and aggregation induced by interaction potentials.
    The node representations obtained through aggregation-diffusion equations exhibit metastability, indicating that features can aggregate into multiple clusters. In addition, the dynamics within these clusters can persist for long time periods, offering the potential to alleviate over-smoothing effects. 
    This nonlinear diffusion in our model generalizes existing diffusion-based models and establishes a connection with classical GNNs. We prove that GRADE achieves competitive performance across various benchmarks and alleviates the over-smoothing issue in GNNs evidenced by the enhanced Dirichlet energy.
\end{abstract}


\section{Introduction}\label{sec.1}
\IEEEPARstart{I}{n} recent years, graph deep learning methods have emerged as one of the most important approaches for handling structured data, achieving impressive results in core scientific problems such as protein and drug design \cite{jing2021learning,dauparas2022robust,atz2021geometric}, materials discovery \cite{merchant2023scaling}, combinatorial optimization \cite{cappart2023combinatorial,schuetz2022combinatorial}. Typical graph deep learning method is Graph Neural Networks (GNNs) \cite{kipf2017semi,velivckovic2018graph}, which construct models based on message passing\cite{gilmer2017neural}. In addition to GNNs, researchers have been exploring more architectures for graph deep learning \cite{xhonneux2020continuous}. Since the proposal of neural ordinary differential equations (Neural ODEs) \cite{chen2018neural}, DE-based continuous neural networks \cite{ruthotto2020deep,kong2020sde} have made significant progress, allowing people to understand and design such models from a dynamical systems perspective \cite{weinan2017proposal,li2022deep}. 

There is a close connection between message passing and graph diffusion equations, leading to the development of numerous continuous graph neural networks based on diffusion \cite{chamberlain2021grand,chamberlain2021blend,thorpe2022grand++,wu2023difformer,wang2023acmp,choi2023gread}. However, the diffusion process exhibits a smoothing effect. The second law of thermodynamics states that an isolated system naturally diffuses an equilibrium state. Thus,  the diffusion in DE-based GNNs can lead to the over-smoothing phenomenon observed in classical GNNs, where node representations become overly consistent and lose expressive power \cite{LiHW18,oono2020graph}. Moreover, taking into account the role of nonlinear activation functions in the expressive power of neural networks \cite{hornik1991approximation,devore2021neural}, considering solely linear graph diffusion may result in a loss of model expressive power.

Aggregation-diffusion equations \cite{carrillo2019aggregation} is mean-field limit of interacting particles driven by interactions and repulsion. Typically, these equations consist of nonlinear diffusion terms and interaction terms. They are capable of modeling the aggregation-diffusion behavior of cells, populations, human societies and etc. Intuitively, tasks on real graph data such as social networks naturally require the aggregation effects, which are missing in the design of diffusion-based continuous GNNs. Due to the interaction terms, there exists a balance between and diffusion, resulting in the system being in a metastable state. Metastability~\cite{mascia2013metastability} has been studied in biological systems to describe the evolution of macroscopic systems. In systems exhibiting metastability, the dynamics tend to settle into a local equilibrium state characterized by multiple clusters, rather than converging to a global equilibrium at an exponential rate as observed in diffusion processes. This phenomenon is particularly relevant in the context of diffusion-based continuous Graph Neural Networks (GNNs), where it contributes to the over-smoothing.

\begin{figure*}[t]
    \centering
\includegraphics[width=0.9\linewidth]{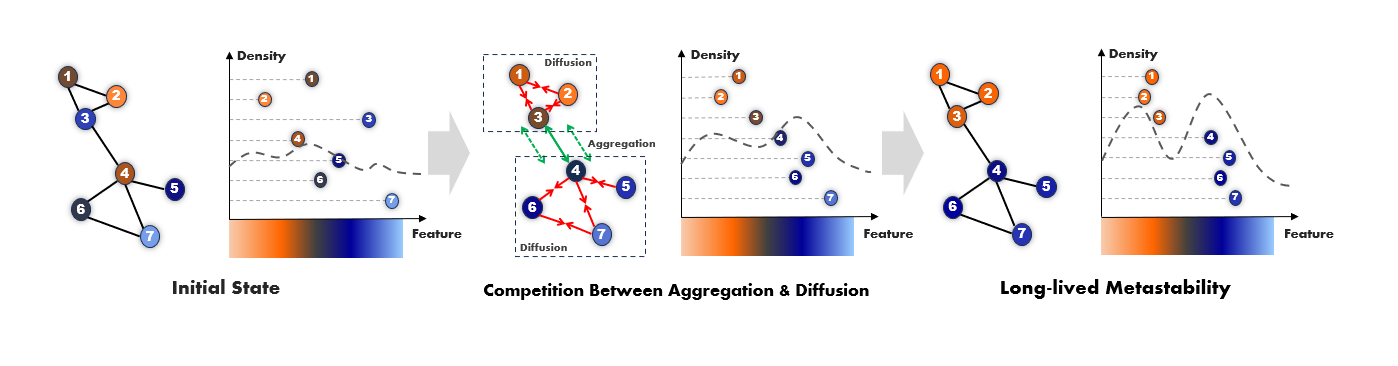}
    \vspace{-2em}
    \caption{An illustration for evalution of GRADE. 
    Due to the appearance of metastable behavior in nonlinear aggregation-diffusion equations, graph features $x(t)$ 
   may be  clustered and contained in several disconnected components.
      Although the metastable states will eventually transition to the ultimate equilibrium state, the system still requires a large amount of time lingering in each metastable state.
    }
    \label{fig:frameworks1}
\end{figure*}

We propose GRADE based on aggregation-diffusion equations. The dynamics of the GRADE consist of nonlinear diffusion and interaction potentials, with flexible options for both diffusion and interaction kernels. Due to metastability, Aggregation-diffusion equations have clustered solutions \cite{carrillo2019aggregation}, which means that node representations driven by Aggregation-diffusion equations will exhibit clustering patterns, thereby mitigating the over-smoothing issue. For typical logarithmic potentials, we prove that GRADE does not suffer from over-smoothing. Additionally, 
existing diffusion-based continuous GNNs can be viewed as linear degenerated versions of our model. We also establish the connection between the GRADE and classical GNNs.

Our contributions can be summarized as follows:
\begin{enumerate}

	\item We extend aggregation-diffusion equations to the graph setting, which can be seen as deriving equations describing the microscopic motion of nodes from equations describing the macroscopic evolution of a system.
	\item We propose the GRADE based on aggregation-diffusion equations on graphs. We prove GRADE can mitigate over-smoothing due to the metastable effect. 
	\item We provide experimental evidence to support the competitive performance of GRADE across various benchmarks. Moreover, we show its capability to prevent Dirichlet energy from vanishing, thereby alleviating the over-smoothing challenge in GNNs.
\end{enumerate}

\section{Background of Graph Neural Networks}\label{sec.2}
\textbf{Notations. } Let $\mathcal{G}=(\mathcal{V},\mathcal{E})$ be a graph with $n$ nodes and $m$ edges, denoted as $|\mathcal{V}|=n$ and $|\mathcal{E}|=m$. We use the notation   $(u,v) \in \mathcal{E}$ to represent the edge between $u$ and $v$, $\mathcal{N}(u)$ to denote the neighbor of node $u$. $ x_u(t)\in\mathbb{R}^{d} $ be the feature (state) of node $ u $. Graph connectivity is represented using an adjacency matrix $W=(w_{u v}) \in\{0,1\}^{n\times n}$, where $w_{u v}=1 $ if $(u,v) \in \mathcal{E}$ and $w_{u v}=0$ otherwise. The graph Laplacian is defined as $L=I- D^{-\frac{1}{2}} W D^{-\frac{1}{2}}$, where $D $ is the diagonal matrix of node degrees and $D_{u u}=d_{u}$ refers to the degree of node $u$.

\textbf{Classical Graph Neural Networks. }  Graph convolutional networks (GCNs) \cite{kipf2017semi} capture node representations with convolutional operators. Specifically, the convolutional layer is expressed as: 
$$
X^{k+1}= \tilde{D}^{- \frac{1}{2}} \tilde{W} \tilde{D}  ^{-\frac{1}{2}}X^{k} \Theta,
$$
where $\tilde{W}=W+I$, and $\tilde{D}$ is the corresponding degree matrix related to the graph adding self-loop. Written in node form, we have:
\begin{equation}\label{eq.GCN}
    x_u^{k+1} =\sum_{v \in \mathcal{V}} \tilde{a}(u,v) x_v^k \Theta,
\end{equation}
where $\tilde{a}(u,v)=\frac{\tilde{w}(u,v)}{\sqrt{\tilde{d}_u \tilde{d}_v}}$ is the node weight. 

Graph attention networks (GATs) \cite{velivckovic2018graph} is another classical graph neural network adding the attention mechanism to GNNs. Substituting the node weights in Eq.\ref{eq.GCN} with the following attention weights: 
\begin{equation}\label{coeff.GAT}
            \alpha_{u v}^k=\frac{\left.\exp \left(\sigma(a^{(k) T} \left[\Theta x_u^k \| \Theta x_v^k\right] \right)\right)}{\sum_{l \in \mathcal{N}_u} \exp \left(\sigma(a^{(k) T} \left[\Theta x_u^k \| \Theta x_l^k\right])\right)}
\end{equation}
leads to the message passing layer of GAT. By assigning attention weights to different nodes, GAT highlights the most relevant neighbors, improving the nodes representation.

\textbf{Continuous Graph Neural Networks. }
    Continuous-time Graph neural networks are a rapidly growing research field of graph representation learning \cite{wu2023difformer, BehmaneshKO23}. Inspired by Neural ODEs\cite{chen2018neural}, many recent works \cite{chamberlain2021grand, thorpe2022grand++, wang2023acmp} formulate the massage-passing propagation in terms of the graph dynamical systems and parameterize the derivative of hidden states instead of modeling a discrete sequence of hidden layers. 
    Then the forward process is expressed as the ordinary differential equation (ODE) initial value problem:
    \begin{equation}\label{eq.int}
        X(T)=X(0)+ \int_{0}^{T} f(X(t),t) dt,
    \end{equation}
    where $X(0)$ is the initial value and the solution $X(T) $ serves the output layer. $f(X(t),t)=\frac{\partial X(t)}{\partial t}$ is modeled by a neural network. Many ODE solvers have been proposed to solve this integral problem such as the Euler method and the Runge-Kutta method.
    GRAND \cite{chamberlain2021grand} leverages graph diffusion equation to formulate the continuous GNNS in the spirit of Neural ODEs. 
    Specifically, the graph diffusion equation in GRAND is expressed as: 
    \begin{equation}\label{eq.GRAND}
        \frac{\partial X(t)}{\partial t}:= [A(X(t))-I]X(t),
    \end{equation}
    where $A(X(t))=(a(X(u,t),X(v,t)))_{u,v \in \mathcal{V}}$ is the attention matrix. 
    
    Recently, \cite{choi2023gread} propose a reaction diffusion equation-based GNN (GREAD) that considers the most popular types of reaction equations expressed as:
    \begin{equation}
         \frac{\partial X(t)}{\partial t}:= \alpha [A(X(t))-I]X(t) + \beta r(X(t)),
    \end{equation}
    where $r(X(t))$ is called the reaction term. Moreover, GREAD incorporates seven types of reaction equations, serving as a comprehensive framework of reaction diffusion based GNNs. In this work, we extend the prior work by using the aggregating-diffusion equation.
    \\

\textbf{Over-smoothing problem in GNNs} poses a significant challenge in the graph representation learning. \

Specifically, over-smoothing refers to the phenomenon that node features convergence to the uniform limit, which can be quantitatively analyzed using Dirichlet energy \cite{wang2023acmp, rusch2022gcon, rusch2022gradient}, as shown in Definition \ref{df.dirichlet_energy}.

\section{A Brief Review of Metastability}\label{sec:review}
\textbf{The metastable states.} Metastability is a widespread phenomenon that occurs in a diverse array of systems, including physics of disordered systems, chemical reaction networks, population biology,
 neural circuits and so on. For example, in chemistry, the mixing of two reactive compounds, such as oxygen and hydrogen, may result in a metastable state that can persist for a very long time. However, when triggered by a spark, it transitions very rapidly to the stable state (water).

\textbf{Characterizing metastable states precisely.}
 In 1971, Penrose and Lebowitz proposed a framework for a rigorous theory of metastability for particle systemsin statistical mechanics, where for the first time the dynamical and static aspects of the problem were coupled in a precise way. They characterised metastable states via three conditions\cite{penrose1971rigorous}:
\begin{enumerate}
\item Only one thermodynamic phase is present;
		\item The lifetime of the metastable state is very long, which means a system that starts in this state is likely to take a long time to get out;
	\item  
Once the system has gotten out, it is unlikely to return. 
	\end{enumerate}

The distinguishing feature of a metastable state is that, the transition from the metastable state to the stable state is an “irreversible” process, wherein the return time is much longer than the decay time.

\textbf{Phase transition. } 
From the point of view of statistical mechanics, metastability \cite{bovier2016metastability} is the dynamical signature of a first-order phase transition. Given a probability distribution $P$  evolves according to  the Fokker-Planck equation, as the example constructed in  \cite{biroli2001metastable}, the systems admits strictly two ergodic components,
corresponding to twofold ground state degeneracy of the
Fokker-Planck operator. 

The phenomenon discussed above can be described by the words in probability. 
Given a Markov process $X$ with state space $S$ and discrete or continuous time. Let $P$ denote the law of $X$ and $P_{x}$ the law of $X$ conditioned on $X_{0}=x$. For $D\subset S$, let $\tau_{D}$ denote the first hitting time of $X$ in $D$
\begin{align*}
\tau_{D}=\inf\{t>0:X(t)\in D\}.
\end{align*}
An intuitively appealing definition of metastability could be the following\cite{bovier2016metastability}:
A family of Markov processes is called metastable if there exists a collection of disjoint sets $B_{i}\subset S$, $i\in I$ such that
\begin{align*}
\frac{\sup_{x\notin \bigcup_{i\in I} B_{i}}E_{x}[\tau_{\bigcup_{i\in I}}B_{i}]}{\inf_{i\in I}\inf_{x\in B_{i}}E_{x}[\tau_{\bigcup_{j\in I\backslash i}}B_{j}]}=o(1),
\end{align*}
 where $o(1)$ should be thought of as a small intrinsic parameter that characterises the “degree” of metastability.

\textbf{Long time of metastability.} 
In physical systems, metastability typically refers to the long-lived occupation of a state with higher energy than the lowest energy state. For instance, metastable states in van der Waals-Maxwell theory\cite{penrose1971rigorous}, and  spin systems like Ising model\cite{capocaccia1974study} or spin glass models\cite{biroli2001metastable}.

 The long time spent in the metastable state is is attributed to the relaxation process of the metastable state towards the equilibrium state, which necessitates overcoming a effective energy barriers\cite{cheng1998role} that hinder the system from easily transitioning to lower energy states. This state is very sensitive to perturbations, even external perturbations, whether spontaneous or induced, can lead the system to escape the metastable state eventually. 
 
 Although the metastable state will reach the ultimate equilibrium state with lowest energy sooner or later,
the metastable dynamics may linger for a potentially asymptotically long time in each metastable state.

\section{GRADE with Metastability}\label{sec.3}
In this section, we first derive Aggregation-diffusion equations on the graph and discuss its metastability. Next, we propose our model GRADE, as shown in Fig. \ref{fig:frameworks1}, where node features can be updated using neural ODE solver.
\subsection{Aggregation-diffusion Equations on Graphs}\label{sec.3.1}

Aggregation-diffusion equations describe the evolution of macroscopic systems, which evaluates the balance between the attraction and the repulsion. It can be formulated as 
\begin{equation}\label{AD_eq}
	\frac{\partial X}{\partial t}=\Delta\sigma(X)+\nabla\cdot(X\nabla(\kappa*X)),
	\end{equation}
	where $ \sigma $ is a monotonically increasing nonlinear function, $  \kappa:\mathbb{R}^{d}\rightarrow\mathbb{R} $ is interaction kernel and $ * $ denotes convolution. 
 
 From a microscopic view, we derive the aggregation-diffusion equation for the state of each node on a discrete graph. For a fixed node $u \in \mathcal{V}$, graph convolution with the interaction kernel $\kappa$ performing on the state $x_u(t)$ is
	\begin{equation}\label{conv}
	\kappa* x_u(t)=\sum_{v\in\mathcal{N}(u)}\kappa[x_u(t)-x_v(t)]x_v(t).
	\end{equation}

Given $\kappa_{u,v} = \kappa[x_u(t) - x_v(t)]$, the gradient of $\kappa * x(u,t)$ with respect to the neighbors $v \in \mathcal{N}(u)$ of node $u$ is
		$$[\nabla\kappa* x]_{u,v}=\sum_{k\in\mathcal{N}(v)}\kappa_{v,k}x_k(t)-\sum_{l\in\mathcal{N}(u)}\kappa_{u,l}x_l(t).$$

The divergence on the graph is defined as $[\nabla\cdot\mathscr{X}]_{u}:=\sum_{(u,v)\in\mathcal{E}}\mathscr{X}_{(u,v)}$, where $\mathscr{X}_{(u,v)}$ denotes the state corresponding to the edge $(u,v)$. Then,
	\begin{align*}\
 &\nabla\cdot[x_u(t)\odot\nabla(\kappa*x_u(t))]=\\&\sum_{v\in\mathcal{N}(u)}x_u(t)\odot[\sum_{k\in\mathcal{N}(v)}\kappa_{v,k}x_k(t)-\sum_{l\in\mathcal{N}(u)}\kappa_{u,l}x_l(t)],
	\end{align*}
	where $ \odot $ denote the Hadamard product. Eventually, Aggregation-diffusion equations on graphs is
	\begin{equation}\label{GAD_eq}
	\begin{aligned}
	&\frac{\mathrm{d} x_u(t)}{\mathrm{d} t} =\sum_{v\in\mathcal{V}}w_{u,v}[\sigma(x_v(t))-\sigma(x_u(t))]\\&+\sum_{v\in\mathcal{V}}w_{u,v}x_u(t)\odot[\sum_{k\in\mathcal{N}(v)}\kappa_{v,k}x_k(t)-\sum_{l\in\mathcal{N}(u)}\kappa_{u,l}x_l(t)], 
	\end{aligned}
	\end{equation}
	where $ w_{u,v} $ is the entry of adjacency matrix $ W $. 
    
\subsection{Metastability of Aggregation-diffusion Equation}\label{sec.3.2}

As illustrated in the Sec.~\ref{sec:review}, we have discussed how the concept of metastability is applicable to various biological systems, including chemotaxis, animal swarming, and pedestrian movements. These phenomena in mathematical biology are often modeled using a class of partial differential equations featuring interactions and nonlinear diffusion, such as the Patlak–Keller–Segel model or the more general aggregation-diffusion equations in Eq.~\eqref{AD_eq}.

Diffusion processes are typically associated with short-range repulsive forces, characterized by the active function $\sigma$. Conversely, particle interactions are often modeled using long-range attractive interaction kernels \cite{fagioli2018solutions}. In this scenario, attractive and repulsive forces compete \cite{hoffmann2017keller}, resulting in complex long-term behavior of solutions and yielding a diverse array of equilibrium configurations. The specific nature of these configurations depends on factors such as the non-linear active function $\sigma$, the chosen interaction potential $\kappa$, the interaction strength, and the dimensionality of the system.

While ongoing research in theory continues to explore various aspects, many questions, including those surrounding metastability, remain unanswered. Numerical simulations suggest that metastable behavior emerges in aggregation-diffusion equations when interactions are recognized by a finite radius of perception \cite{carrillo2019aggregation}.

Specifically, when dynamics are governed by local interactions, solutions can form clustered patterns, and their long-term behavior often leads to the emergence of one or more clusters. Interestingly, the number of stationary clusters can be directly inferred from the multiplicity of leading spectral eigenvalues of certain stochastic matrices \cite{motsch2014heterophilious}. In other words, for certain interaction potentials with compact support, the steady-state solutions may comprise several disconnected components. These findings, along with related research, bear resemblance to metastable states observed in physics, particularly in spin glass systems as discussed in Section~\ref{sec:review}.

Metastable behavior has been observed in systems with interaction kernels that are unbounded away from the origin \cite{carrillo2019aggregation}, such as the repulsive-attractive interaction potential described by power laws, \textit{e.g.}, $W(x) = x^{4} - x^{2}$ in one dimension. This type of interaction potential has inspired the development of a new message passing mechanism called ACMP \cite{wang2023acmp} in graph neural networks, which effectively mitigates over-smoothing. Viewing this phenomenon from a physical perspective, metastable behaviors offer valuable insights for addressing over-smoothing issues in GNNs.

\subsection{Our model}\label{sec.3.3}

\textbf{GRADE. } 
Given an input graph $\mathcal{G}$ and node features $x_u^{\text{IN}}$, $u\in\mathcal{V}$, GRADE first generates embeddings using an encoder layer. This encoder can be a simple multi-layer perceptron with learnable parameters:
	$$ x_u(0)=\text{Encoder}(x_{u}^{\text{IN}}). $$
Next is the Aggregation-diffusion layer, which is a Neural ODE module
\begin{equation}\label{eq.neural_ode_module}
	 x_{u}(T)=x_{u}(0)+\int_{0}^{T}f(x_{u}(t))\mathrm{d} t, 
    \end{equation}
	where the dynamic of $f$ is based on aggregation-diffusion equations on the graph (\ref{GAD_eq}). Then
	\begin{align*}
		&f(x_{u}(t)):=\frac{\mathrm{d} x_u(t)}{\mathrm{d} t}=\sum_{v\in\mathcal{V}}w_{u,v}[\sigma(x_v(t))-\sigma(x_u(t))]\\&+\sum_{v\in\mathcal{V}}w_{u,v}x_u(t)\odot[\sum_{k\in\mathcal{N}(v)}\kappa_{v,k}x_k(t)-\sum_{l\in\mathcal{N}(u)}\kappa_{u,l}x_l(t)].
	\end{align*}
Written in matrix form:
 \begin{equation} \label{eq.f_matrix}
     f(X(t))=(A-I)\sigma(X(t))+(A-I)X(t)\odot K(X(t))X(t).
 \end{equation}
 
	Finally, the output of the Aggregation-diffusion layer, denoted as $X(T)$, is put into a decoding output layer to obtain the output:
	$$ Y^{\text{OUT}}=\text{Decoder}(X(T)). $$

\textbf{Over-smoothing. } Consider the case that interaction kernel $\kappa$ choosing from a class of singular function such as $\kappa=\log|x|$, then equation~\eqref{GAD_eq}~ can be written as
\begin{equation}\label{over_eq}
	\begin{aligned}
	&\frac{\mathrm{d} x_u(t)}{\mathrm{d} t} =\sum_{v\in\mathcal{V}}w_{u,v}[\sigma(x_v(t))-\sigma(x_u(t))]\\&+\sum_{v\in\mathcal{V}}w_{u,v}x_u(t)\odot[\sum_{k\in\mathcal{N}(v)}\vert x_v(t)-x_k(t)\vert^{-2-\delta}x_k(t)\\&-\sum_{l\in\mathcal{N}(u)}\vert x_u(t)-x_l(t)\vert ^{-2-\delta}x_l(t)], 
	\end{aligned}
\end{equation}	
\begin{definition}\label{df.dirichlet_energy}

Node features are over-smoothing, if there exists a constant vector $c$ such that for any $u\in \mathcal{V}$
\begin{align*}
	\lim_{t\rightarrow+\infty} x_{u}(t)=c.
	\end{align*}
\end{definition}
\begin{theorem}\label{thm3.2}
Suppose that the graph $ \mathcal{G}=(\mathcal{V},\mathcal{E}) $ is connected and there is a node $ u^{*}\in\mathcal{V} $ with degree 1.
Then the feature determined by Aggregation-diffusion equations~\eqref{GAD_eq}~on graph will mitigate over-smoothing.
\end{theorem}

\textbf{Attention adjacency matrix. }
	 By substituting the adjacency matrix from Eq.\eqref{eq.f_matrix}  of GATs \cite{velivckovic2018graph}, we can express GRADE-GAT as:
    \begin{equation*}
        \frac{dX(t)}{dt}= (A-I)\sigma(X(t))+(A-I)X(t)\odot K(X(t))X(t),
    \end{equation*}

    where $A=(a(x_u(t),x_v(t)))$ is the attention adjacency matrix, with the entry of matrix $A$ as follows:
    \begin{equation}\label{eq.attention}
        a(x_u(t),x_v(t))=softmax \left(\frac{(\Theta x_u(t))^T\Theta x_v(t))}{d_{\Theta}} \right).
    \end{equation}
    Here $\Theta$ is the matrix with learnable parameters and $d_{\Theta}$ represent the scale factor. These diffusion coefficients can be learned through background propagation.

\textbf{Attention interaction kernel. }
	The attention mechanism can be introduced to the interaction kernel. In fact, the attention weights can be considered as a kernel, making this process quite straightforward and intuitive. Then Eq. \eqref{conv} is rewritten as follows:
    \begin{equation}
        \kappa*x_u(t) =\sum_{v \in \mathcal{V}} k(x_u,x_v) x_v(t),
    \end{equation}
    where $k(x_u,x_v)$ is the attention weight analogous to Eq. \eqref{eq.attention}. The flexible options for diffusion coefficients and interaction kernel exhibit impressive expressive capabilities.

\textbf{Neural ODE solver. } 
    For numerically implementing the forward propagation and backward propagation of the proposed model, many existing explicit and implicit schemes serve as the ODE solver such as the Euler method, Runge-Kutta method, and Dormand-Prince5 method \cite{chamberlain2021grand, lu2018beyond, chen2018neural}. For instance, to solve the output $X(T)$ from our
    model (\ref{eq.neural_ode_module}), the Euler method leads to the following discrete form:  
    \begin{equation}
        \frac{X^{k+1}-X^k}{\delta}=(A-I)\sigma(X^k)+(A-I)X^k \odot K(X^k)X^k,
    \end{equation}
    where $\delta$ is the time step and $X^k=X(\delta k)$ refers to the node representation obtained in step $k$. Starting from the initial value $X(0)$ and performing the iterative computations, we can obtain the output $X(T)$ at time $T$.

\begin{table*}[t]
\caption{Results on real-world homophilic datasets: The values refer to mean $\pm$ std. dev. accuracy for 10 initialization and 100 random train-val-test splits. We show the best three methods in \BEST{red} (first), \SECOND{blue} (second), and \THIRD{purple} (third).}
\label{tab:Homophilic Result}
\begin{center}
\resizebox{\linewidth}{!}{
\begin{tabular}{l cccccc}
\toprule 
Dataset               & \textbf{Cora} & \textbf{CiteSeer} & \textbf{PubMed} & \textbf{Coauthor CS} & \textbf{Computer} & \textbf{Photo} \\
Homophily level & $\mathbf{0.83}$ & $\mathbf{0.71}$ & $\mathbf{0.79}$& $\mathbf{0.80}$& $\mathbf{0.77}$& $\mathbf{0.83}$ \\ 
\midrule
\textbf{GCN} \cite{kipf2017semi}               & $81.5\pm1.3$      & $71.9\pm1.9$          & $77.8\pm2.9$        & $91.1\pm0.5$             & $82.6\pm2.4$          & $91.2\pm1.2$       \\
\textbf{GAT} \cite{velivckovic2018graph}               & $81.8\pm1.3$      & $71.4\pm1.9$          & $78.7\pm2.3$        & $90.5\pm0.6$             & $78.0\pm19.0$                & $85.7\pm20.3$               \\
\textbf{GAT-ppr} \cite{velivckovic2018graph}           & $81.6\pm0.3$      & $68.5\pm0.2$          & $76.7\pm0.3$        & $91.3\pm0.1$             & $85.4\pm0.1$          & $90.9\pm0.3$       \\
\textbf{MoNet} \cite{Monti2016GeometricDL}             & $81.3\pm1.3$      & $71.2\pm2.0$          & $78.6\pm2.3$        & $90.8\pm0.6$             & $83.5\pm2.2$          & $91.2\pm2.3$       \\
\textbf{GraphSage-mean} \cite{Hamilton2017}    & $79.2\pm7.7$      & $71.6\pm2.0$          & $77.4\pm2.2$        & $91.3\pm2.8$             & $82.4\pm1.8$          & $91.4\pm1.3$       \\
\textbf{GraphSage-max} \cite{Hamilton2017}   & 79.2 $\pm$ 7.7 & 71.6 $\pm$ 1.9 & 77.4 $\pm$ 2.2 & 91.3 $\pm$ 2.8 & 82.4 $\pm$ 1.8 & 91.4 $\pm$ 1.3  \\  \hline
\textbf{CGNN} \cite{xhonneux2020continuous}              & $81.4\pm1.6$      & $66.9\pm1.8$          & $66.6\pm4.4$        & $\THIRD{92.3\pm0.2}$             & $80.3\pm2.0$         & $91.4\pm1.5$      \\
\textbf{GDE} \cite{Poli2019}               & $78.7\pm2.2$      & $71.8\pm1.1$          & $73.9\pm3.7$        & $91.6\pm0.1$             & $81.9\pm0.6$          & $92.4\pm2.0$       \\
\textbf{GRAND-l} \cite{chamberlain2021grand}           & $\THIRD{83.6\pm1.0}$      & $73.4\pm0.5$          & $78.8\pm1.7$        & $\BEST{92.9\pm0.4}$             & $83.7\pm1.2$          & $92.3\pm0.9$   \\   
\textbf{GRAND++} \cite{thorpe2022grand++}           & $82.5 \pm 0.7$      & $\THIRD{73.8 \pm 2.7}$          & $\BEST{79.8 \pm 1.6}$        & $90.2 \pm 0.3$             & $84.1 \pm 0.5$          & $\BEST{93.5 \pm 0.4}$   \\ 
\textbf{ACMP-GAT} \cite{wang2023acmp}       & $82.3 \pm 0.5$     &  $\BEST{75.5 \pm 1.0}$                   & $\THIRD{79.4 \pm 0.4}$      & $91.8 \pm 0.1$  & $84.4 \pm1.6$                   & $91.1\pm0.7$\\
\midrule
\textbf{GRADE-GAT}        & $\BEST{84.2 \pm 0.8}$     &  $73.7 \pm 1.3$                   & $79.3 \pm 0.2$      & $92.1 \pm 0.1$  & $\SECOND{85.4 \pm1.1}$                   & $\SECOND{92.9\pm0.8}$\\
\textbf{GRADE-Gaussian}        & $\SECOND{84.0 \pm 0.7}$     &  $\SECOND{74.7 \pm 1.0}$                   & $78.9 \pm 0.2$      & $\SECOND{92.7 \pm 0.2}$  & $\THIRD{84.6 \pm1.4}$                   & $\THIRD{92.7\pm 0.7}$\\
\textbf{GRADE-log}        & $82.4 \pm 0.9$     &  $73.5 \pm 1.7$                   & $\SECOND{79.5 \pm 0.4}$      & $91.3 \pm 0.4$  & $\BEST{85.7 \pm 1.5}$                   & ${90.3\pm 1.0}$ \\
    \bottomrule
    \end{tabular}
}
    \end{center}

\end{table*}
\section{Related work}\label{sec.4}
\subsection{Continuous-time GNNs} 
Continuous-time neural networks have gained prominence since \cite{chen2018neural} initiated Neural ODEs, which treat neural networks as ordinary differential equations (ODEs) and use ODE solvers to conduct the forward process and backward propagation. Subsequent research has focused on enhancing training methods \cite{li2020scalable, ghosh2020steer, xu2018jknet, zhao2020pairnorm}. Many studies have explored Graph Neural Networks based on diffusion equations on graphs \cite{PDE-GCN, choi2023gread, wang2023acmp}.\cite{chamberlain2021grand} extended Neural ODEs to graph neural networks (GNNs) with the GRAND model, which utilizes graph diffusion equations to model feature-updating dynamics. BLEND \cite{chamberlain2021blend} incorporates the positional feature into the node features and utilizes the Beltrami flow to jointly evolve them. However, pure diffusion processes often lead to the oversmoothing problem \cite{thorpe2022grand++, wang2023acmp, choi2023gread}. To mitigate this issue, one common approach is adding an additional term to prevent Dirichlet energy from decaying to zero \cite{wang2023acmp, rusch2022gcon, rusch2022gradient}. GREAD, proposed by \cite{choi2023gread}, incorporates nearly all commonly used additional terms based on the reaction-diffusion equation.

In this study, we formulate a graph neural network using the aggregation-diffusion equations. These equations incorporate nonlinear terms that significantly enhance the network's expressive capability in representing node features. Unlike previous models that primarily focus on the long-term behavior of Graph Neural Networks, our approach with GRADE leverages metastability to ensure local equilibrium states rather than global oversmoothing. As a result, our model offers a more comprehensive and versatile framework compared to existing diffusion-based GNN models.

\subsection{Classical GNNs} 
Numerous Graph Neural Network (GNN) models have emerged, rooted in the concept of aggregating and updating node-level features, inspired by the Massage Passing Neural Network (MPNN) paradigm \cite{gilmer2017neural}. Classical GNNs like Graph Convolutional Networks (GCN) \cite{kipf2017semi} and Graph Attention Networks (GAT) \cite{velivckovic2018graph} are traditionally implemented using discrete schemes. However, to achieve continuous transformations akin to GCN, recent approaches such as those proposed by \cite{chen2018neural, wang2021dissecting} decouple the parameter transformation and message passing processes. This separation allows for parameter learning to occur solely at the beginning and end of the propagation process, resulting in a continuous transformation formulation for classical GNNs: $X(T)=X(0)+\int_{0}^{T} A(X(t)) X(t) dt$, where $A(X(t))$ denotes the attention matrix. This continuous formulation can be interpreted as non-linear diffusion on the graph.

GRADE integrates both the diffusion process and interaction potentials, encompassing diffusion-based GNNs when the impact of the interaction kernel is disregarded. Thus, through this analysis, we demonstrate that GRADE serves as a general framework that incorporates existing classical GNNs while enabling non-linear diffusion.

\begin{table*}[t]
    \centering
    \caption{Results on real-world heterophilic datasets: The values refer to mean $\pm$ std. dev. accuracy for 10 initialization and 100 random train-val-test splits. We show the best three methods in \BEST{red} (first), \SECOND{blue} (second), and \THIRD{purple} (third).}
    \resizebox{\linewidth}{!}{
    \begin{tabular}{l cccccc}\toprule
        Dataset     & \textbf{Texas}      & \textbf{Wisconsin}  & \textbf{Cornell}    & \textbf{Film}       & \textbf{Squirrel}   & \textbf{Chameleon}  \\
        Homophily level & $\mathbf{0.11}$ & $\mathbf{0.21}$ & $\mathbf{0.30}$& $\mathbf{0.22}$& $\mathbf{0.22}$& $\mathbf{0.23}$ \\ \midrule
        \textbf{Geom-GCN}~\cite{pei2020geomGCN}	& $66.7\pm2.7$ & $64.5\pm3.6$ & $60.5\pm3.6$ & $31.5\pm1.1$ & $38.1\pm0.9 $& $60.0\pm2.8$ \\
        \textbf{H2GCN}	    & $84.8\pm7.2$ & $87.6\pm4.9$ & $82.7\pm5.2$ & $35.7\pm1.0$ & $36.4\pm1.8$ & $60.1\pm2.1$ \\
        
    \textbf{GGCN} \cite{yan2021GGCN} & $84.8\pm4.5$ & $86.8\pm3.2$ & $\SECOND{85.6\pm6.6}$ & $\SECOND{37.5\pm1.5}$ & $55.1\pm1.5$ & $\SECOND{71.1\pm1.8}$ \\
    \textbf{LINKX} \cite{lim2022LINKX} & $74.6\pm8.3$ & $75.4\pm5.7$ & $77.8\pm5.8$ & $36.1\pm1.5$ & $\BEST{61.8\pm1.8}$ & $68.4\pm1.3$ \\
    \textbf{GloGNN} \cite{li2022GloGNN} & $84.3\pm4.1$ & $87.0\pm3.5$ & $83.5\pm4.2$ & $\THIRD{37.3\pm1.3}$ & $\THIRD{57.5\pm1.3}$ & $\THIRD{69.8\pm2.4}$ \\
    \textbf{ACM-GCN} \cite{luan2022ACMGCN} & $87.8\pm4.4$ & $\SECOND{88.4\pm3.2}$ & $85.1\pm6.1$ & $36.3\pm1.1$ & $54.4\pm1.9$ & $66.9\pm1.9$ \\
        \midrule
\textbf{CGNN}~\cite{xhonneux2020continuous} & $71.4\pm4.1$ & $74.3\pm7.3$ & $66.2\pm7.7$ & $35.9\pm0.9$ & $29.2\pm1.1$ & $46.9\pm1.7$ \\
        \textbf{GDE}~\cite{Poli2019} 	    & $74.1\pm6.9$ & $79.8\pm5.6$ & $82.4\pm7.1$ & $35.4\pm1.3$ & $35.9\pm1.9$ & $47.8\pm2.1$\\
        \textbf{GRAND-l}~\cite{chamberlain2021grand}       & $75.7\pm7.3$ & $79.4\pm3.6$ & $82.2\pm7.1$ & $35.1\pm1.2$ & $38.3\pm1.7$ & $52.1\pm3.0$ \\
        \textbf{ACMP-GCN}~\cite{wang2023acmp}       & $\THIRD{86.2}\pm3.0$ & ${86.1\pm4.0}$ & $\THIRD{85.4\pm7.0}$
 & $35.6\pm1.1$ & $40.1\pm1.5$ & $54.7\pm2.5$ \\
        \textbf{GREAD-BS}~\cite{choi2023gread} 
                    & $\BEST{88.9\pm3.7}$ & $\BEST{89.4\pm3.3}$ & $\SECOND{86.5\pm7.2} $
                    & $\BEST{37.9\pm1.2}$
                    & $\SECOND{59.2\pm1.4}$ & $\BEST{71.4\pm1.3}$ \\ 
                    \midrule
        \textbf{GRADE-GAT}        & $\SECOND{88.3\pm3.5}$     &  $\THIRD{87.7\pm 3.7}$                   & ${83.3 \pm 7.0}$      & $36.8 \pm 1.1$  & $45.4 \pm1.9$                   & $55.1\pm 2.4$\\
     \bottomrule 
    \end{tabular}}
    \label{tab:heter result}
\end{table*}

\section{Experiments}\label{sec.5}

This section presents a comprehensive analysis of the proposed GRADE model through extensive experimentation. Initially, we compare GRADE against various baselines in the context of node classification tasks. Subsequently, we leverage Dirichlet Energy to analyze the capability of GRADE to tackle the over-smoothing challenge.

Our implementation adheres closely to the methodology outlined in the official codebase of GRAND~\cite{chamberlain2021grand}, ensuring a fair and rigorous comparison. Specifically, we maintain consistency in hyper-parameter settings and the choice of numerical differential equation solvers with GRAND. All experiments are performed on a server equipped with eight NVIDIA Tesla V100 graphics cards.

\subsection{Node Classification}
We conduct a comparative analysis of GRADE against several state-of-the-art GNN  architectures across a diverse set of node classification benchmarks. 

\paragraph{Datasets.}
The benchmarks utilized in our study cover datasets exhibiting both homophilic and heterophilic characteristics. Homophilic graph data is characterized by nodes with similar attributes tending to connect together, while heterophilic graph data exhibits a low level of homophily, indicating that most neighbors do not share the same labels as the source nodes.
To comprehensively evaluate the performance, we employ a variety of real-world datasets. These include three widely studied citation graphs: Cora~\cite{mccallum2000automating}, Citeseer~\cite{sen2008collective}, and Pubmed~\cite{namata2012query}. Additionally, we utilize the coauthor graph CoauthorCS~\cite{shchur2018pitfalls}, as well as the Amazon co-purchasing graphs Computer and Photo~\cite{mcauley2015image}, selecting the largest connected component in each case.
Furthermore, we include three heterophilic datasets characterized by low homophily ratios: Texas, Wisconsin, and Cornell from the WebKB dataset\footnote{http://www.cs.cmu.edu/afs/cs.cmu.edu/project/theo-11/www/wwkb/}, and Chameleon, Squirrel~\cite{benedek2021musae}, Film~\cite{tang2009social},

\paragraph{Baselines} We incorporate a comprehensive set of baseline models that have been widely compared in prior studies~\cite{chamberlain2021grand,thorpe2022grand++,choi2023gread,wang2023acmp}. These baselines can be categorized into the following third groups:
\begin{compactenum}
    \item Classical GNN methods: GCN~\cite{kipf2017semi}, GAT~\cite{velivckovic2018graph}, GraphSAGE~\cite{Hamilton2017}, and MoNet~\cite{Monti2016GeometricDL}.
    \item GNN methods for heterophilic tasks: 
    Geom-GCN~\cite{pei2020geomGCN}, H2GCN~\cite{zhu2020h2gcn}, 
    GPR-GNN~\cite{chien2021GPRGNN},
    GGCN~\cite{yan2021GGCN}, LINKX~\cite{lim2022LINKX}, GloGNN~\cite{li2022GloGNN} and ACM-GCN~\cite{luan2022ACMGCN}.
    \item Continuous GNN methods: GDE~\cite{Poli2019}, CGNN~\cite{xhonneux2020continuous}, GRAND~\cite{chamberlain2021grand}, 
    ACMP~\cite{wang2023acmp}, 
    GREAD~\cite{choi2023gread}.
\end{compactenum}

In all experiments, we conduct 100 train-validation-test splits for each dataset, employing 10 random seeds for each split. Following the works by\cite{wang2023acmp} and\cite{choi2023gread}, we report both the mean and standard deviation of accuracy. When the accuracy of a baseline model is known and its experimental conditions match ours, we utilize the officially announced accuracy. However, if the accuracy is not readily available, we execute the baseline model using its official codes and employ their suggested hyper-parameter ranges.

\paragraph{Results on homophilic dataset.} Table~\ref{tab:Homophilic Result} presents the detailed classification performance on homophilic datasets. Our method GRADE demonstrates competitive performance across all cases. Notably, traditional GNNs exhibit a linear relationship between the number of parameters and depth. In contrast, continuous GNNs share parameters across layers due to the utilization of time-independent weights, resulting in significantly fewer parameters compared to competing methods. For instance, versions of GCN and GAT require approximately 150K and 1.6M parameters, respectively, while GRAND and our model require fewer than 100K parameters. However, it is observed that continuous GNNs outperform traditional GNNs in this task due to their deeper continuous layers. Additionally, compared to GRAND, our method achieves superior performance in most cases, attributed to the introduction of the aggregation term. This proves the aggregation-diffusion equation facilitates the Graph Neural Diffusion.

\paragraph{Results on different kernels.} In our investigation, we examined the efficacy of three alternative kernel types: one based on attention, one based on logarithms, and another based on the Gaussian kernel. To ensure stability during training, we employed normalization techniques to guarantee that the sum of kernels equaled unity.
As depicted in Table~\ref{tab:Homophilic Result}, all three kernels exhibited commendable performance. The results also underscored the robust performance of both the attention and Gaussian kernels across diverse datasets. However, the log kernel encountered challenges in certain datasets, primarily due to its singularity issue, leading to suboptimal results. Consequently, in practical applications, we prioritize the utilization of the attention kernel due to its superior stability and consistent performance across a wide range of datasets. 

\paragraph{Results on heterophilic dataset.} 
We assess the performance of GRADE on heterophilic graphs, which pose greater challenges compared to homophilic datasets. As demonstrated in Table~\ref{tab:heter result}, GRADE exhibits competitive performance on this task when compared with GRAND. This highlights the effectiveness of introducing the aggregation term in enhancing the performance of GNNs on heterophilic datasets.
Moreover, GRADE shows comparable performance to another state-of-the-art model, ACMP, and slightly lags behind GREAD. This difference can be attributed to GREAD's utilization of more parameter search techniques, whereas we adhered to the parameters of GRAND.

\begin{figure}
    \centering
\includegraphics[width=1\linewidth]{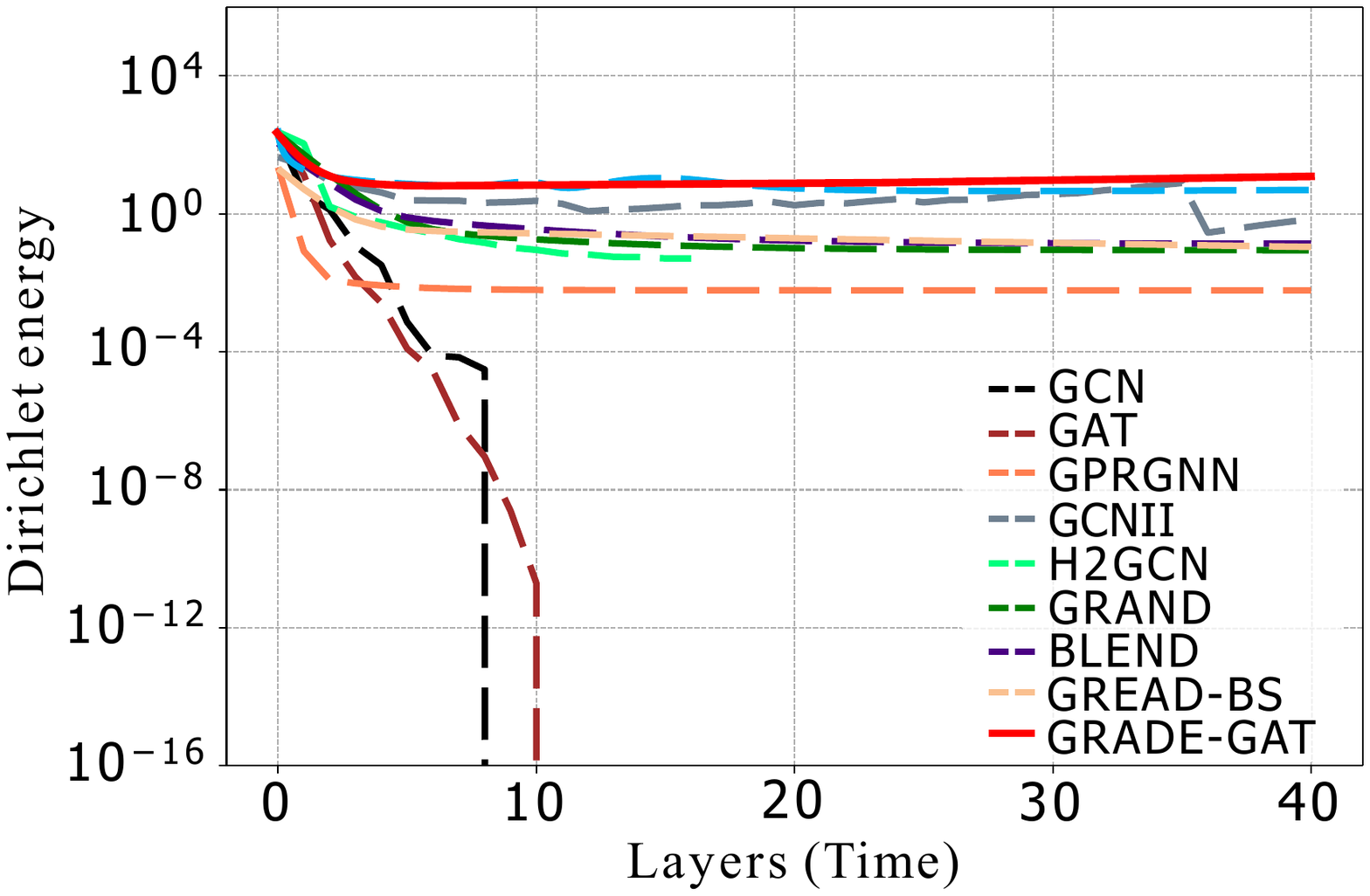}
    \vspace{-2em}
    \caption{Dirichlet energy on the synthetic random graph.}
    \label{fig:d-energy}
\end{figure}
\subsection{Dirichlet Energy}
To analyze the degree of over-smoothing~\cite{oono2020graph}, we follow prior works~\cite{wang2023acmp,choi2023gread} to evaluate Dirichlet energy~\cite{rusch2023survey} on the synthetic dataset called cSBMs~\cite{Deshpande2018cSBM}, which comprises an undirected graph with 100 nodes in a two-dimensional space, where two classes are randomly connected with a probability of $p = 0.9$.  In our analysis, we consider GNNs comprising 40 layers, and we compute the layer-wise Dirichlet energy define on the hidden features:
\begin{align*}
	E(X)=&\frac{1}{N} \sum_{i\in\mathcal{V}}\sum_{j\in\mathcal{N}(i)}w_{i,j}\Vert x_{i}-x_{j}\Vert^{2}. 
	\end{align*}
In alignment with convention, we set $w_{i,j} = 1$ if two nodes $i$ and $j$ are adjacent, and $w_{i,j} = 0$ otherwise. In addition to baseline methods, we also include some approaches designed to alleviate over-smoothing, such as Blend~\cite{chamberlain2021blend}, GPRGNN~\cite{chien2021GPRGNN}, and GCNII~\cite{chen2020gcnii}.

As illustrated in Figure~\ref{fig:d-energy}, traditional Graph Neural Networks (GNNs) such as GCN and GAT exhibit obvious over-smoothing, characterized by the exponential decay of Dirichlet energy to zero. Conversely, the Dirichlet energy of other methods remains bounded over time. Notably, among these methods, GRADE demonstrates the most effective mitigation of over-smoothing, attributed to the inclusion of the aggregation term as supported by our theoretical analysis. On the other hand, GRAND, which relies solely on a diffusion term, continues to face over-smoothing challenges despite being equipped with only a learnable diffusion term.

\section{Conclusions}\label{sec.6}
This paper presents a class of continuous Graph neural network grounded in graph aggregation-diffusion equations, called GRADE. By offering a flexible choice for both non-linear diffusion and non-local interaction potentials, GRADE serves as one of the most generalized frameworks. 
The metastability in GRADE ensures the local equilibrium while a pure diffusion process exhibits the over-smoothing effect. 
Our experiments show competitive performance across diverse benchmarks. Additionally,
GRADE shows an impressive capacity to preserve Dirichlet energy, thereby offering substantial evidence for its effectiveness in addressing the over-smoothing problem.

\section*{Potential Broader Impact}
This paper presents advancements in the field of Graph Neural Networks by incorporating aggregation-diffusion equations. We consider our work has no potential negative impact on society or the community.

\newpage


\newpage
\onecolumn
{\appendix[Proof]
\begin{proof}
[proof of Theorem \ref{thm3.2}]
If for any $ u\in\mathcal{V} $, we have $x_{u}(t)\rightarrow c>0$ as $t\rightarrow+\infty$. Then for arbitrary $\varepsilon>0$, there exists a $T$ large enough such that
$|x_{u}-x_{v}|<\varepsilon$ and 
$|x_{u}|<\varepsilon+c$. 
For $\log$ kernel, consider the node $u^{*}$ with neighbor $v^{*}$ satisfies
\begin{align*}
    \frac{\mathrm{d} x_{u^{*}}(t)}{\mathrm{d} t} &=\sum_{v\in\mathcal{V}}w_{u^{*},v}[\sigma(x_v(t))-\sigma(x_{u^{*}}(t))]\\&+\sum_{v\in\mathcal{V}}w_{u^{*},v}x_{u^{*}}(t)\odot\left[\sum_{k\in\mathcal{N}(v)}\log[\vert x_v(t)-x_k(t)\vert] x_k(t)-\sum_{l\in\mathcal{N}(u^{*})}\log[\vert x_{u^{*}}(t)-x_l(t)\vert] x_l(t)\right]\\
    &\leq \sum_{v\in\mathcal{V}}w_{u^{*},v}Lip_{\sigma}\vert x_v(t)-x_{u^{*}}(t)\vert\\&+\sum_{v\in\mathcal{V}}w_{u^{*},v}x_{u^{*}}(t)\odot[\sum_{k\in\mathcal{N}(v)}\log[\vert x_v(t)-x_k(t)\vert] x_k(t)-\log[\vert x_{u^{*}}(t)-x_{v^{*}}(t)\vert]x_{v^{*}}(t)]
\end{align*}
Because $v^{*}$ is the unique neighbor of $u^{*}$, then  for $t\geq T$
\begin{align*}
&\sum_{v\in\mathcal{V}}w_{u^{*},v}x_{u^{*}}(t)\odot[\sum_{k\in\mathcal{N}(v)}\log[\vert x_v(t)-x_k(t)\vert] x_k(t)-\log[\vert x_{u^{*}}(t)-x_{v^{*}}(t)\vert]x_{v^{*}}(t)]\\
    =&w_{u^{*},v^{*}}x_{u^{*}}(t)\odot[\sum_{k\in\mathcal{N}(v^{*})\backslash u^{*}}\log[\vert x_{v^{*}}(t)-x_k(t)\vert]x_k(t)+(x_{u^{*}}(t)-x_{v^{*}}(t))\log[\vert x_{u^{*}}(t)-x_{v^{*}}(t)\vert]]\\
    \leq &(c-\varepsilon)^{2}(\deg(v^{*})-1)\log[\varepsilon]-(c+\varepsilon)\varepsilon\log[\varepsilon],
\end{align*}
which means
$$\frac{\mathrm{d} x_{u^{*}}(t)}{\mathrm{d} t} \leq Lip_{\sigma}\varepsilon+(c-\varepsilon)^{2}(\deg(v^{*})-1)\log[\varepsilon]-(c+\varepsilon)\varepsilon\log[\varepsilon].$$

Then we can take $\varepsilon>0$ small enough such that for $t\geq T$
$$\frac{\mathrm{d} x_{u^{*}}(t)}{\mathrm{d} t} \leq Lip_{\sigma}\varepsilon+(c-\varepsilon)^{2}(\deg(v^{*})-1)\log[\varepsilon]-(c+\varepsilon)\varepsilon\log[\varepsilon]\leq -1.$$
Finally,
\begin{align*}
  x_{u^{*}}(t) &\leq x_{u^{*}}(T)-t.
\end{align*}
This contradicts the assumption $c>0$.

\end{proof}
\vfill

\end{document}